\pdfoutput=1

\documentclass[11pt]{article}

\usepackage[]{EMNLP2023}

\usepackage{times}
\usepackage{latexsym}

\usepackage[T1]{fontenc}

\usepackage[utf8]{inputenc}

\usepackage{microtype}

\usepackage{inconsolata}

\usepackage{soul}
\usepackage{booktabs}
\usepackage{amsmath,caption}
\usepackage{colortbl}
\usepackage{arydshln}
\usepackage{multirow}
\usepackage{multicol}
\usepackage{hyperref}

\usepackage{xspace}
\usepackage{adjustbox}
\usepackage{subcaption}
\usepackage{longtable}
\usepackage{makecell}
\usepackage{tabularx}
\usepackage{pifont}
\usepackage{colortbl}
\usepackage{array}
\usepackage{verbatim}
\usepackage{svg}
\usepackage{hyperref}

\newcommand\hybridqa{{HybridQA}\xspace}
\newcommand\tapex{\textsc{TaPEx}\xspace}

\newcommand\tacube{\textsc{TaCube}\xspace}
\newcommand\ourdata{\textsc{\emph{KET-QA}}\xspace}

\newcommand\kns{\emph{kNS}\xspace}
\newcommand\retriever{\emph{MKBR}\xspace}
\newcommand\rk{R@$k$\xspace}
\newcommand\recallk{Recall@$k$\xspace}
\newcommand\tblgray{\rowcolor{gray!20}}

\title{KET-QA: A Dataset for Knowledge Enhanced Table Question Answering}

\author{
~Mengkang Hu\textsuperscript{1},
~Haoyu Dong\textsuperscript{2}$\thanks{~~~Corresponding author.}$, 
~Ping Luo\textsuperscript{1}, 
~Shi Han\textsuperscript{2}, 
~Dongmei Zhang\textsuperscript{2}\\
 ~\textsuperscript{1} The University of Hong Kong, 
 ~\textsuperscript{2}Microsoft \\
mkhu@connect.hku.hk, pluo.lhi@gmail.com,
\{hadong, shihan, dongmeiz\}@microsoft.com}

\begin{document}
\maketitle
\begin{abstract}
Due to the concise and structured nature of tables, the knowledge contained therein may be incomplete or missing, posing a significant challenge for table question answering (TableQA) and data analysis systems. 
Most existing datasets either fail to address the issue of external knowledge in TableQA or only utilize unstructured text as supplementary information for tables. 
In this paper, we propose to use a knowledge base (KB) as the external knowledge source for TableQA and construct a dataset \ourdata with fine-grained gold evidence annotation.
Each table in the dataset corresponds to a sub-graph of the entire KB, and every question requires the integration of information from both the table and the sub-graph to be answered. 
To extract pertinent information from the vast knowledge sub-graph and apply it to TableQA, we design a retriever-reasoner structured pipeline model.
Experimental results demonstrate that our model consistently achieves remarkable relative performance improvements ranging from 1.9 to 6.5 times and absolute improvements of 11.66\% to 44.64\% on EM scores across three distinct settings (fine-tuning, zero-shot, and few-shot), in comparison with solely relying on table information in the traditional TableQA manner.
However, even the best model achieves a 60.23\% EM score, which still lags behind the human-level performance, highlighting the challenging nature of \ourdata for the question-answering community.
We also provide a human evaluation of error cases to analyze further the aspects in which the model can be improved. Project page: \href{https://ketqa.github.io/}{https://ketqa.github.io/}.
\end{abstract}

\section{Introduction}
\label{sec:intro}
\begin{figure*}[t]
\vspace{-0.5cm}
    \begin{center}
    \includegraphics[scale=0.32]{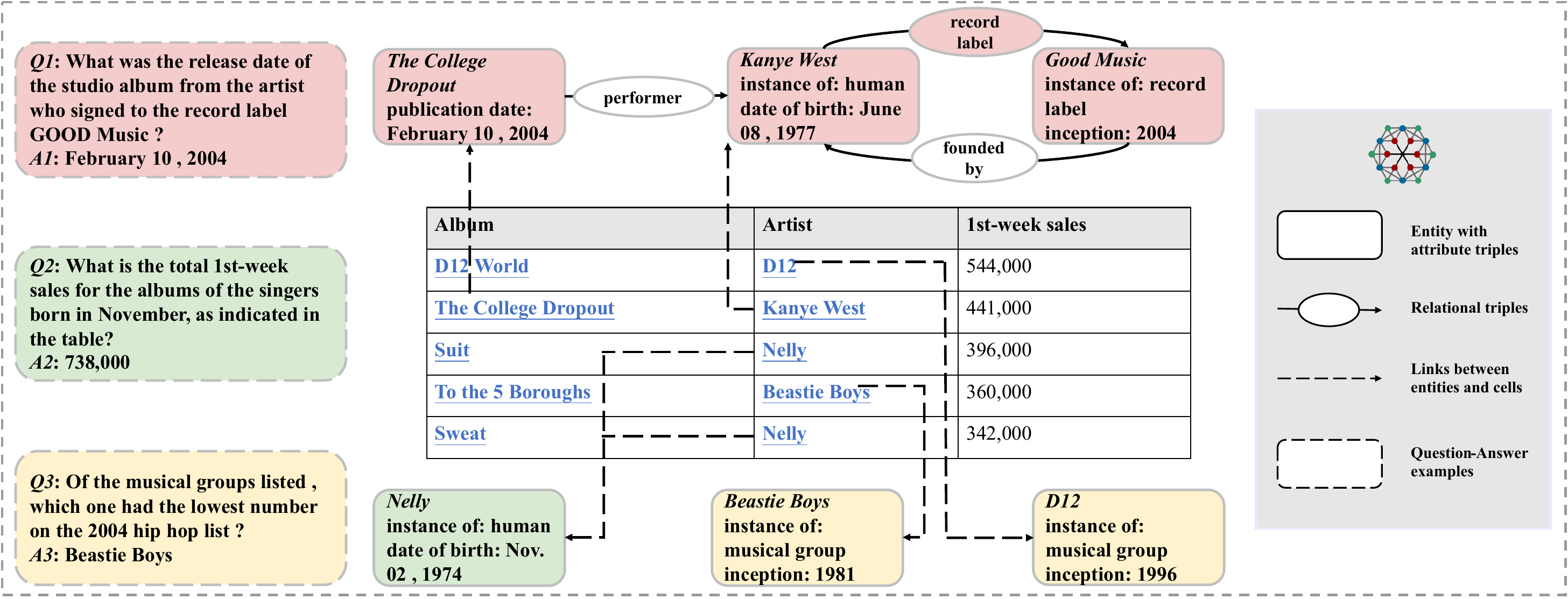}
    \end{center}
\caption{Overview of \ourdata. Only a partial table and knowledge graph are displayed for better visualization. The examples of red, green, and yellow respectively represent three distinct sources of answers: \textit{In-KB}, \textit{Calculated}, and \textit{In-Table}. Corresponding gold evidence for each question is highlighted using the respective colors. Each underlined cell is linked to an entity in Wikidata.}
\label{fig:dataset_overview}
\vspace{-0.5cm}
\end{figure*}

As a kind of distinct information source, the table is extensively researched for the task of table question answering (TableQA) with numerous practical applications~\cite{pasupat2015compositional, yu2018spider, chen-etal-2020-hybridqa}.
Its objective is to answer questions by utilizing specific tables as context. 
However, owing to the inherent conciseness and organized structure of tables, the information they contain may be incomplete or absent. Consequently, humans, as well as question-answering (QA) systems, may necessitate background knowledge to acquire comprehensive information.
These questions that are difficult to answer due to missing information in the table have sparked research interest in addressing the need for external knowledge~\cite{chen-etal-2020-hybridqa, cheng2023binding}. As is explored in \tacube~\cite{zhou-etal-2022-tacube}, approximately 10\% of the samples in WTQ~\cite{pasupat2015compositional} belong to this category.
We define \textit{External Knowledge} as factual information required to answer a given question beyond what is provided in the table. 
For example, in Figure~\ref{fig:dataset_overview}, to answer question 1 \textit{"What was the release date of the studio album from the artist who signed to the record label GOOD Music?"}, a QA system needs to know not only the record label to which each artist belongs in the table but also the release dates of each album, both of which are missing from the table. 
We consider handling external knowledge required samples as a significant challenge for current TableQA systems.

As shown in Table~\ref{tab:comparison}, most existing TableQA datasets do not explicitly emphasize the inclusion of external knowledge required questions, although annotators may introduce their prior knowledge during the annotation process. 
While some efforts have been made to incorporate textual information as external knowledge~\cite{chen-etal-2020-hybridqa, zhu2021tat, chen-etal-2021-finqa}, there has been a significant oversight in leveraging knowledge graphs, which are widely recognized as an equally prevalent knowledge source.
To address this gap, we propose \ourdata, in which each table is associated with a sub-graph from the Wikidata knowledge base~\cite{DennyVrandecic2014WikidataAF}, serving as supplementary information for question answering. Each question in \ourdata requires external knowledge to answer, meaning that it necessitates the integration of the table and knowledge base.
To construct \ourdata, we face two main challenges: 
\textbf{(i)} Identifying tables that can be well augmented with an external knowledge base is difficult. We find that there exists an inherent mapping relationship between Wikipedia pages and Wikidata entities~\cite{DennyVrandecic2014WikidataAF}, and cells in Wikipedia tables are well linked with Wikipedia pages. These factors naturally connect Wikipedia tables to Wikidata. 
\textbf{(ii)} Proposing natural external knowledge required questions is labor-intensive. 
Alternatively, we choose to re-annotate natural human-created questions in \hybridqa, which leverage unstructured passages in Wikipedia as external knowledge source.
We start by substituting the external knowledge source in \hybridqa with Wikidata, and cells that originally corresponded to Wikipedia pages are now replaced with entities from Wikidata. Then, we extract a one-hop sub-graph for each table with its corresponding entities.
Subsequently, we employ a two-stage annotation approach to re-annotate the question-answer pairs.
In the first stage, annotators assess whether a sample is self-contained. That is, to answer the question, it is necessary to incorporate information from the knowledge base, and no additional data is required.
In the second stage, annotators annotate the fine-grained gold evidence necessary to answer the given question from the sub-graph.
Finally, we collected 9,421 questions and 5,721 tables. Each table corresponds to a sub-graph with 1,696.7 triples on average.
Also, we believe that the annotation of fine-grained gold evidence for each question can facilitate a more in-depth exploration of external knowledge required samples within the TableQA research community.
Examples from \ourdata are depicted in Figure~\ref{fig:dataset_overview}.

Incorporating a knowledge base into the TableQA process poses two challenges: 
\textbf{(i)} The amount of information contained within the grounded sub-graph remains substantial and redundant for a specific question; 
\textbf{(ii)} Integrating three different types of data, namely questions (unstructured text), tables (semi-structured), and a knowledge base (structured), for reasoning purposes.
To address the aforementioned challenges, we devise a retriever-reasoner pipeline model, which consists of two steps: first, retrieving relevant triples from the sub-graph based on the given question, and then utilizing a pre-trained language model (PLM) to incorporate the question, table, and retrieved triples, ultimately producing the final answer. This process is illustrated in Figure~\ref{fig:model-overview}.
Benefiting from the fine-grained annotations of gold evidence in \ourdata, we primarily focus on optimizing the retriever in terms of both time efficiency and performance. The resulting \textbf{M}ultistage \textbf{KB} \textbf{R}etriever (\retriever) not only demonstrated state-of-the-art performance  with an 83.47 Recall@20 score but also achieved a remarkable speed improvement of 9.3 times.
In the final question-answering experiments, incorporating the information from the knowledge graph led to relative performance enhancements ranging from 1.9 to 6.5 times and absolute improvements of 11.66\% to 44.64\% in terms of EM scores across different models and settings (fine-tuning, few-shot, zero-shot), as compared to solely utilizing table information for answering questions.
Moreover, we conducted a comprehensive comparison with two additional external sources of knowledge (LLM-generated knowledge and unstructured text). This extensive comparison further confirmed the benefits and advantages of utilizing a knowledge graph.
Despite the substantial performance boost achieved by incorporating the information stored in the knowledge graph into TableQA, its performance still falls short in comparison to human-level performance, with the highest 60.23\% EM score. Consequently, we consider \ourdata to be a challenging problem for the question-answering community. To analyze further the bottlenecks of the current model, we conducted a manual error analysis, revealing potential areas for further improvement in both the retriever and the reasoner.


\section{Task Definition}
\label{sec:task_definition}
A table $T$ is a structured arrangement of data that is organized into rows, i.e., $T = \{c_{ij}\} = \{r_i\}$, where $i$ and $j$ represent the coordinates of rows and columns, respectively.
A knowledge base $ G $ (KB) is a collection of factual information that is formalized as a set of statements that can be categorized into two types, i.e., $G = \{(e_1, p, e_2)\} \cup \{(e, a, v)\}$. The first type represents relational triples, where entities $e_1$ and $e_2$ are related by a relation $p$. The second type represents attribute triples, where an entity $e$ has an attribute $a$ with a value $v$. The set of all possible entities, relations, and attributes is denoted by $E$, $R$, and $A$, respectively. 
In \ourdata, a cell in the table may be linked to some entities in the KB. For example, each underlined cell in Figure~\ref{fig:dataset_overview} will be linked to an entity in Wikidata~\cite{DennyVrandecic2014WikidataAF}. We denote this relation between cells and entities as function $f$.
$E' = \bigcup f(c_{ij})$ represents the union of entities corresponding to all cells in the table. We retrieve a sub-graph by taking $E'$ and their one-hop neighbors. That is, each table corresponds to a sub-graph $ \mathcal{G} $ from the entire KB, which serves as supplementary information for TableQA.

The process of \ourdata is as follows: given a table $ T $, the grounded knowledge sub-graph $ \mathcal{G} $, and a natural language question $ q $, output $a$ that answers the question according to the context. 
The source of $a$ could be divided into three categories: \textbf{(i)} \textit{In-KB}: an entity or attribute value in the knowledge base; \textbf{(ii)} \textit{In-Table}: a cell in the table; \textbf{(iii)} \textit{Calculated}: a value that was calculated on the table and knowledge base by using numerical operations like 'count', 'sum', 'difference', etc. Example inputs and outputs are shown in Figure~\ref{fig:dataset_overview}.

\section{Dataset Construction}


\subsection{Table Collection} 
\label{sec:table-collection}
Each hyperlink in Wikipedia tables points to a Wikipedia page. According to the data model of Wikidata~\cite{DennyVrandecic2014WikidataAF}
there exists alignments between Wikipedia pages and Wikidata entities.
Therefore, in order to incorporate more external knowledge, our collected table should include as many hyperlinks as possible.
With a similar purpose, \hybridqa~\cite{chen-etal-2020-hybridqa} has collected 13,000 tables, and 35\% of the total cells have hyperlinks, thus we took the collection as the table set.
Then, we mapped the hyperlinked Wikipedia pages to Wikidata entities by creating an index from a Wikipedia SQL dump with WikiMapper.\footnote{\url{https://github.com/jcklie/wikimapper}} Finally, each table is mapped to 44.3 entities in Wikidata on average.


\subsection{Knowledge Base Construction} 
Following the definition in Section~\ref{sec:task_definition}, for each table, we took all linked entities and extracted a one-hop sub-graph of Wikidata. \footnote{The data was collected in September 2022 via \url{https://www.wikidata.org/w/api.php}}
Then, we performed post-processing on the obtained knowledge sub-graph, discarding attributes such as "globe-coordinate" and "URL" that are not likely to be used.

\subsection{Question/Answer Annotation} 
\label{sec: question-answer-annotation}
The collection of question-answer pairs is built on \hybridqa, where each question requires the integration of information from both Wikipedia passages and tables to be answered. By simply removing the Wikipedia passages, these questions become naturally categorized as external knowledge required samples as defined in Section \ref{sec:intro}.
We started by filtering \hybridqa via simple heuristics and obtained 23,600 question-answer pairs that are most likely to meet the definition of \ourdata. The rules and details of filtering can be found in Appendix~\ref{appendix:dataset-details-qa-filtering}.
Then we hired 43 graduate crowd-sourced workers to manually re-annotate the question-answer pairs and 2 experienced experts to perform regular inspections and fix labeling errors.
The annotation process consists of two stages. 
In the first stage, annotators are required to determine whether one given question-answer pair is suitable for \ourdata setting. The criteria for this stage are: \textbf{(i)} The question cannot be answered solely by relying on one of the information sources; \textbf{(ii)} Annotated results must be self-contained, which means no additional information sources will be required to answer questions other than knowledge graphs and tables.
In the second stage, annotators are tasked with annotating the fine-grained gold evidence required to answer a specific question from the sub-graph corresponding to the current table. 
One item of gold evidence should include two components: \textbf{(i)} the question-relevant triple; \textbf{(ii)} the corresponding cell. For convenience, we represent it as $((i, j), t)$, where $i$ and $j$ denote the row and column coordinates of the cell (starting from zero), and $t$ represents a triple. Each question-answer pair would possess multiple pieces of gold evidence. Take the first question in Figure~\ref{fig:dataset_overview} as an example, the gold evidence for this instance would be \{((2, 1), (Kanye West, record label, Good Music)), ((2, 0), (The College Dropout, publication date, February 10, 2004))\}. 
Additionally, when there is a mismatch in the answer format or issues arise with the original annotations, the annotators will modify the question-answer pair.

\subsection{Final Review and Quality Control}
Prior to the final review, we assessed the agreement among annotators by collecting and comparing annotations on a random set of 245 question-answer pairs. The results indicated "almost perfect agreement" with Fleiss Kappa~\cite{10.2307/2529310} scores of 0.89 for the first annotation stage and 0.82 for the second annotation stage.
Next, the two experienced experts conducted a final review of all annotations, ensuring that any errors or inconsistencies in the annotations were corrected. 
Then, we applied several rules to filter low-quality annotations, which can be found in Appendix~\ref{appendix:low-quality-filtering}. 
Finally, we obtained 9,421 high-quality annotations and 
we randomly split the data into training/development/test sets with 80\%/10\%/10\% respectively.

\subsection{Dataset Analysis}

\begin{table}[bthp]
  \centering
  \scriptsize
  \begin{tabular}{lccccc}
    \toprule
    \multirow{2}[4]{*}{\textbf{Dataset}} & \multicolumn{2}{c}{\textbf{Size}} & \multicolumn{3}{c}{\textbf{External Knowledge}} \\
    \cmidrule{2-6}
    & \textbf{\#Ques.} & \textbf{\#Tables} & \textbf{Type} & \textbf{Source} & \textbf{GE} \\
    \midrule
    WTQ & 22,033 & 2,108 & - & - & - \\
    WikiSQL & 80,654 & 26,521 & - & - & - \\
    Spider & 10,181 & 1,020 & - & - & - \\
    HiTab & 10,672 & 3,597 & - & - & - \\
    FeTaQA & 10,330 & 10,330 & - & - & - \\
    HybridQA & 69,611 & 13,000 & Text & Wikipedia & No \\
    TAT-QA & 16,552 & 2,757 & Text & Financial reports & Yes \\
    FinQA & 8,281 & 2,776 & Text & Financial reports & Yes \\
    \tblgray
    KET-QA & 9,421 & 5,721 & KB & Wikidata & Yes \\
    \bottomrule
  \end{tabular}%
  \caption{Comparison between \ourdata and exsisting datasets. \#Ques. and \#Table are the numbers of questions and tables, respectively. GE stands for the annotation of \textbf{G}old \textbf{E}vidence. Datasets: WTQ~\cite{pasupat2015compositional}, WikiSQL~\cite{DBLP:journals/corr/abs-1709-00103}, Spider~\cite{yu2018spider}, HiTab~\cite{cheng-etal-2022-hitab}, FeTaQA~\cite{nan-etal-2022-fetaqa}, HybridQA~\cite{chen-etal-2020-hybridqa}, TAT-QA~\cite{zhu2021tat}, FinQA~\cite{chen-etal-2021-finqa}. 
  }
  \label{tab:comparison}%
\vspace{-0.4cm}
\end{table}

\noindent \textbf{Basic Statistics}
Table~\ref{tab:comparison} shows a comparison of \ourdata with existing table question answering datasets, and Table~\ref{tab:statistics-ket-qa} shows comprehensive statistics. 
Despite not being one of the largest datasets, \ourdata still has several advantages:
\textbf{(i)} It is the first TableQA dataset that utilizes a knowledge base as an external knowledge source;
\textbf{(ii)} It provides alignment between Wikipedia tables and Wikidata entities;
\textbf{(iii)} It includes fine-grained gold evidence annotations from external knowledge sources, enabling more in-depth analyses.

\noindent \textbf{Question Types}
We also analyzed the question types comprehensively and visualized them in Appendix~\ref{sec:question-types-other-datasets} using the heuristic method proposed by~\citet{yang-etal-2018-hotpotqa}. It is noteworthy that, compared to the original \hybridqa, there is a higher proportion of \textit{Who} questions in \ourdata, accounting for 12.6\% as opposed to 9.8\% in \hybridqa. \textit{Who} questions are typically associated with human entities in Wikidata. 

\begin{table}[htbp]
  \centering
  \scriptsize
    \begin{tabular}{ccc}
    \toprule
    \#Words/Ques. & \#Words/Answer & \#Rows/Table \\
    \midrule
    17.2 & 3.3  & 15.8 \\
    \midrule
    \#Columns/Table & \#Entities/Table & \#Triples/Table \\
    \midrule
    4.5  & 41.9 & 1696.7 \\
    \midrule
    Answer in KB & Answer in Table & Calculated Answer \\
    \midrule
     5197 & 4131 & 93 \\
    \bottomrule
    \end{tabular}%
  \caption{Core statistics of \ourdata.
  }
  \label{tab:statistics-ket-qa}%
\vspace{-0.3cm}
\end{table}%

\section{Model}

We propose a retriever-reasoner pipeline model to address the challenges as discussed in Section \ref{sec:intro}.
The overview of this model can be found in Figure~\ref{fig:model-overview}, and 
the detailed probabilistic formalization of this architecture is presented in Appendix~\ref{appendix:probabilistic-formalization}.


\begin{figure}[thbp]
    \centering
    \includegraphics[width=0.45\textwidth]{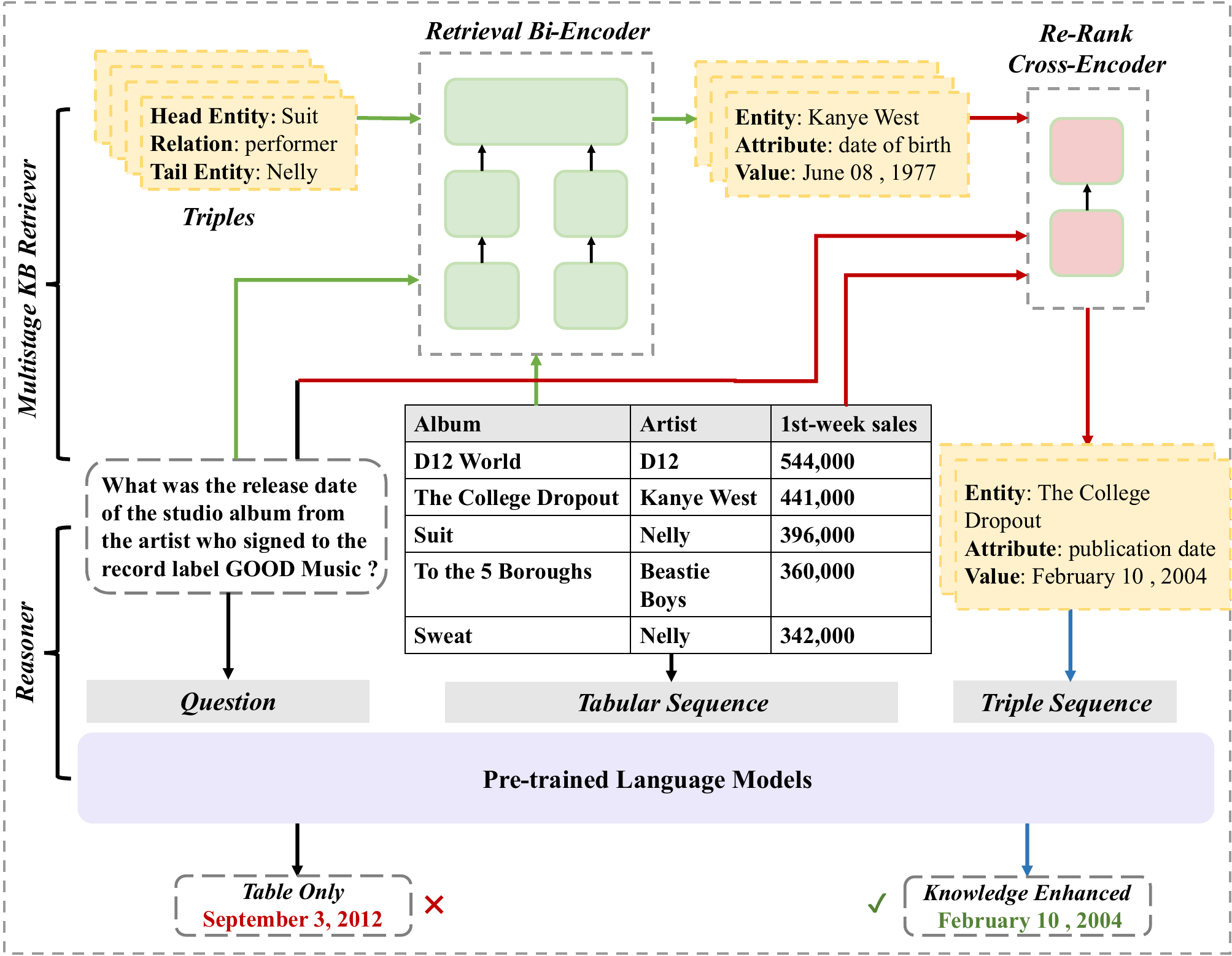}
\caption{Overview of the retriever-reasoner model.}
\label{fig:model-overview}
\vspace{-0.4cm}
\end{figure}

\subsection{Preliminary}

\noindent \textbf{Triple Serialization} To enable PLMs to handle structured information from a knowledge base, we devised a straightforward approach to transform triples {\small $t_1 = (e_1, r, e_2)$} and {\small $t_2=(e, a, v)$} into textual sequence {\small $t^*_1 =  \texttt{\small [HEAD]}, \texttt{\small $\ell(e_1)$},
     \texttt{\small [REL]}, \texttt{\small $\ell(r)$},
     \texttt{\small [TAIL]}, \texttt{\small $\ell(e_2)$}$ }
and {\small $t^*_2 =  \texttt{\small [HEAD]}, \texttt{\small $\ell(e)$},
     \texttt{\small [REL]}, \texttt{\small $\ell(a)$},
     \texttt{\small [TAIL]}, v$}.
Here, \texttt{\small [HEAD]}, \texttt{\small [RELATION]}, \texttt{\small [TAIL]} are special tokens representing distinct components of a triple. The function $\ell(e)$ retrieves the label of $e$ from KB, with the same functionality for $r$ and $a$.

\noindent \textbf{Table Serialization}
With a similar purpose, we adopt the same serialization method in \tapex~\cite{liu2021tapex} to flatten a table $T$ into a tabular sequence $T^*$. 

\subsection{Multistage Knowledge Base Retriever} 
\label{sec:model-retriever}

\begin{figure}[htb]
    \centering       
    \includegraphics[width=0.35\textwidth]{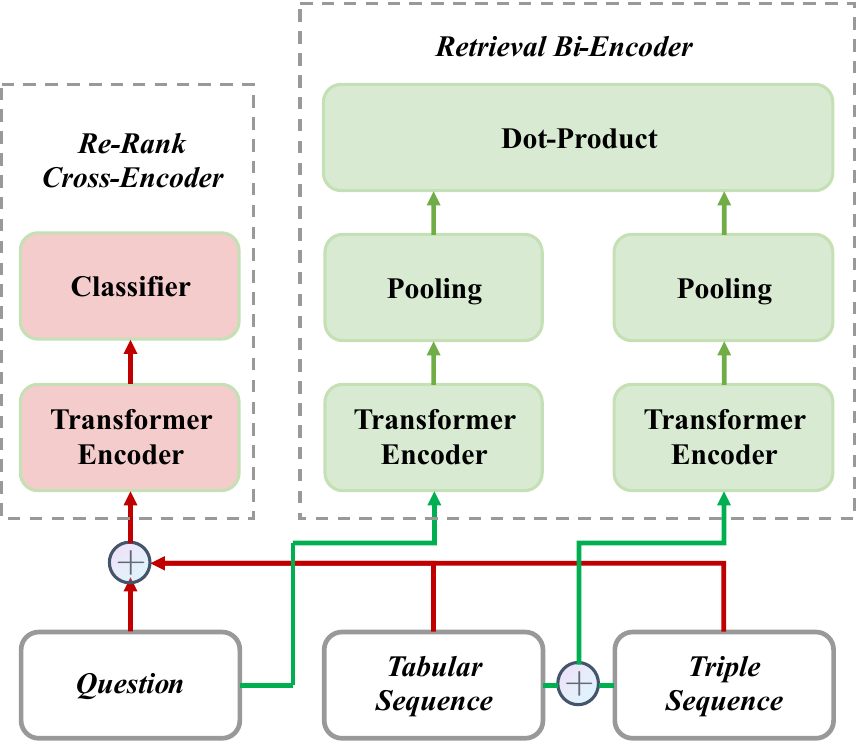}
    \caption{Diagrams of two retrievers. The symbol $\oplus$ denotes the concatenation between text sequences.}
    \label{fig:retriever-architecture}
\end{figure}

\noindent \textbf{Overview} 
The retriever predicts the relevance scores $s(t, q, T)$, and then the top $ k $ triples with the highest score will be fed into the reasoner. 
The retriever we designed draws inspiration from the field of text semantic matching, among which bi-encoder and cross-encoder are two commonly employed model architectures, as is shown in Figure~\ref{fig:retriever-architecture}. 
Normally, cross-encoders can achieve superior performance due to their fine-grained cross-attention inside the PLM. However, in practical TableQA scenarios, there is a significant demand for real-time retrieval of relevant knowledge from the knowledge base. This imposes a substantial requirement on the speed of the retrieval operation. 
On the other hand, a bi-encoder is generally faster and more efficient since it only requires one pass through the input sequence.
Therefore, we propose \textbf{M}ultistage \textbf{KB} \textbf{R}etriever (\retriever), which first utilizes a Retrieval Bi-Encoder to retrieve the top $N$ triple candidates and subsequently employ a Re-Ranker Cross-Encoder for more precise scoring. According to our experimental results, this approach outperforms the standalone usage of either architecture of the retriever while maintaining high run-time efficiency.

\noindent \textbf{Retrieval Dataset}
The $i$-th instance contains one question $q_i$ , one table $T$ , $m$ relevant (positive) triples $t^+_{i, j}$ and $n$ irrelevant (negative) triples $t^-_{i,j}$.
The positive triples are annotated manually as is discussed in Section~\ref{sec: question-answer-annotation}. The negative triples are sampled from non-positive triples within the sub-graph $\mathcal{G}$ since considering all negatives would result in an unbearable computational cost. Rather than simply sampling uniformly, we develop a strategy called \textbf{k}NN \textbf{N}egative \textbf{S}ampling (\kns). 
Firstly, both the question and all triples are encoded into vectors with a pre-trained sentence embedding model. Then, we take $n$ non-positive triples that are closest to the question in the vector space as negatives. \kns aims to choose informative negative samples, which is also studied in~\cite{DBLP:conf/iclr/RobinsonCSJ21, kumar-etal-2019-improving, zhang-stratos-2021-understanding, xiong2021approximate}.

\vspace{-0.2cm}

\noindent \textbf{Retrieval Bi-Encoder}
consists of two modules: a question encoder $E_q$ and a context encoder $E_c$. The concatenation of the serialized table and triple form the context. $E_q$ and $E_c$ are two independent BERT-style transformer networks~\cite{vaswani2017attention}, and we take the representation at the \texttt{\small [CLS]} token as the output vector. 
We define the relevance score using the dot product of the two vectors, i.e., {\small $s(t, q, T) = E_q(q)^\top E_c(T^* \oplus t^*)$}. Where $\oplus$ is the concatenation operator. The bi-encoder is trained using a contrastive loss, detailed formulation can be found in Appendix~\ref{appendix:training-bi-encoder}.

\noindent \textbf{Re-Ranker Cross-Encoder}
directly takes the concatenation of question $q$, serialized table $T^*$, and serialized triple $t^*$ as a joint input to the PLM and generate a relevance score ranging from 0 to 1, i.e., {\small $s(t, q, T) = E(q \oplus T^* \oplus t^*)$}. Specifically, we take the output logits as the relevance score. The training process of the cross-encoder is modeled as a binary classification problem. Positive triples and negative triples are assigned as 1 and 0, respectively.

\noindent \textbf{Triple-Related Sub-Table} The information within the entire table may be redundant for retrieval and could exceed the length limitations of the transformer model. We assume that each row in the table is independent, and the relevance score of a triple $t$ is determined solely by the rows that have a mapping relationship with the head entity. Therefore, we propose to improve the performance of the retriever via extracting a triple-related sub-table $ \mathcal{T} = \{r_i \in T\; | \; \exists \; c_{ij} \in r_i, e \in f(c_{ij})\}$ and take $\mathcal{T}$ as the input of retriever. 
In Section~\ref{sec:retrieval-ablation-study}, we showcase the efficacy of this approach.

\subsection{Reasoner} 
\label{sec:model-reasoner}
The reasoner aims to answer the question with the information from the table and the retrieved triples. For this phase, we follow the trend of directly generating answers using auto-regressive PLMs in question answering area 
~\cite{raffel2020t5, lewis-etal-2020-bart}. 
Meanwhile, PLMs are also powerful for fusing and reasoning on heterogeneous data~\cite{zhou-etal-2022-tacube, chen2022large}. 
Specifically, we concatenate the retrieved information with serialized table and question as the input sequence of PLMs and take the output as the final answer, i.e., $ a' = E_{r}(q \oplus T^* \oplus t^*_0 \oplus \cdots \oplus t^*_k)$.
Where $a'$ is the predicted answer. $E_r$ stands for the reasoner model. $T^*$ is the serialized table and $\{t^*_0, \cdots t^*_k\}$ is the set of serialized triples obtained from the retriever.

\section{Experiments: Evidence Retrieval}

\subsection{Experimental Setup}
\noindent \textbf{Evaluation Metrics}
We introduce a modified version of \recallk(\rk) to evaluate the retrieval performance in the context of \ourdata. 
The purpose of \rk is to measure the percentage of items of gold evidence that are retrieved by the retriever: 
\vspace{-0.2cm}
\begin{equation}
{\small
\begin{aligned}
R@k = \frac{1}{N} \sum_{i=1}^{N} \frac{|{evidence \ retrieved}|_{i}}{|{gold \ evidence}|_{i}}
\end{aligned}}
\vspace{-0.2cm}
\nonumber
\label{eq:recall_at_k}
\end{equation}
In this equation, the numerator counts the relevant items retrieved up to the $k$-th position for the $i$-th instance, while the denominator represents the total number of relevant items for the $i$-th instance.

\noindent \textbf{Baseline Methods} (i) \textit{String Match}: Triples are retrieved based on whether the label of the $r$ or the $e_2$  for relational triples and $a$ or $v$ for attribute triples matches the words in the question; (ii) \textit{Bi-Encoder} and \textit{Cross-Encoder} are used to compare the performance of a single retriever with \retriever.

\noindent \textbf{Implementation Details}
During the training process, 
we applied \kns for the negative sample selection of Bi-Encoder with $n=25$, but random sampling for Cross-Encoder with $n=50$. 
We chose not to apply \kns to the Cross-Encoder because we observed that training it using \kns is highly time-consuming, and the model tends to overfit. 
The hyper-parameter search for $n$ and more training details can be found in Appendix~\ref{appendix:retriever-details}.
The optimal \retriever model is obtained by selecting the best Retrieval Bi-Encoder and Re-Rank Cross-Encoder separately, based on their performance on the development set.
During the inference process, we first retrieve the top 200 triple candidates with the bi-encoder and re-rank them with the cross-encoder.
For all experiments, we report the R@K of different methods on the test set of \ourdata. 
validation performance for each reported test result is reported in Appendix~\ref{sec:validation-results}.

\subsection{Main Results}

\begin{table}[htbp]
\centering
\small 
\begin{tabular}{lcccc} 

\toprule
\textbf{Method} & \multicolumn{1}{c}{\textbf{Top-1}} & \multicolumn{1}{c}{\textbf{Top-5}} & \multicolumn{1}{c}{\textbf{Top-20}} & \multicolumn{1}{c}{\textbf{Top-100}} \\
\midrule
Random & 0.05 & 0.27 & 2.94 & 12.49 \\
String Match & 5.87 & 14.65 & 28.24 & 43.66 \\
Cross-Encoder & 37.83 & 63.84 & 82.14 & \textbf{94.44} \\
Bi-Encoder & 29.17 & 51.95 & 72.12 & 89.62 \\
\tblgray
\retriever & \textbf{38.77} & \textbf{66.04} & \textbf{83.47} & 93.51 \\
\bottomrule
\end{tabular}
\caption{Comparison between retrieval methods on \ourdata test set using \rk ($k \in \{1, 5, 20, 100\}$). 
}
\label{tab:retriever-main-results}
\vspace{-0.3cm}
\end{table}

From Table~\ref{tab:retriever-main-results}, we can derive the conclusion that in scenarios where k is small ($k \leq 20$), \retriever consistently outperforms any single retriever model. We attribute this performance improvement primarily to the complementary nature of the bi-encoder and cross-encoder. The Retrieval Bi-Encoder aids the Re-Rank Cross-Encoder in filtering out a subset of triples that are difficult to distinguish, thus enhancing the overall performance.


\subsection{Ablation Study}
\label{sec:retrieval-ablation-study}
In Section~\ref{sec:model-retriever}, we propose to utilize the triple-related sub-table as the final input for the retriever. 
However, there are two other straightforward approaches for table representation:
\textbf{(i)} Full Table: taking the full table as the input for the retriever;
\textbf{(ii)} No Table: not including the table as input.
As is shown in Table~\ref{tab:ablation-study-table-representation}, the table representation method using the triple-related sub-table is superior to the other two approaches. We believe this is because such a representation preserves the relevant information in the table that can aid in retrieval while minimizing redundant information.

\begin{table}[htbp]
  \centering
  \small
    \begin{tabular}{ccccc}
    \toprule
    \textbf{Table Rep.} & \textbf{Top-1} & \textbf{Top-5} & \textbf{Top-20} & \textbf{Top-100} \\
    \midrule
    \multicolumn{5}{l}{\textit{Bi-Encoder}} \\
    FT    & 25.81 & 49.89 & 71.75 & 88.91 \\
    NT    & 24.05 & 48.67 & 71.89 & 89.13 \\
    \tblgray
    TT    & \textbf{29.17} & \textbf{51.95} & \textbf{72.12} & \textbf{89.62} \\
    \midrule
    \multicolumn{5}{l}{\textit{Cross-Encoder}} \\
    FT    & 28.59 & 58.94 & 78.32 & 94.42 \\
    NT    & 12.27 & 34.41 & 59.78 & 85.04 \\
    \tblgray
    TT    & \textbf{37.83} & \textbf{63.84} & \textbf{82.14} & \textbf{94.44} \\
    \bottomrule
    \end{tabular}%
    \caption{Ablation study on different table representation methods, which can be chosen from \{FT (\textbf{F}ull \textbf{T}able), NT (\textbf{N}o \textbf{T}able), TT (\textbf{T}riple-Related Sub-\textbf{T}able)\}.}
  \label{tab:ablation-study-table-representation}%
\vspace{-0.3cm}
\end{table}

\subsection{Run-time Efficiency}
We conducted run-time tests for \retriever on a remote server equipped with four 16G V100 GPUs.
If we exclusively utilize a Cross-Encoder, the retrieval would take 4.92 seconds per question. However, by incorporating \retriever, the retrieval process is optimized to 0.53 seconds while achieving higher \rk.
On the other hand, the Retrieval Bi-Encoder requires a longer time for the offline generation of knowledge base embeddings. It takes approximately 27.9 seconds per table (with each table corresponding to around 1.7k triples).
%

\section{Experiments: Question Answering}
\subsection{Experimental Setup}
\noindent \textbf{Evaluation Metrics} We applied two widely-used metrics in the question-answering area: exact match (EM) and F1. Detailed introduction of evaluation metrics can be found in Appendix~\ref{appendix:evaluation-metrics-qa}.

\noindent \textbf{Baseline Methods} We take table-only models as baselines to explore whether the question can be answered based solely on the table information in the traditional TableQA manner.

\noindent \textbf{Implementation Details}
We selected \tapex, T5, BART, GPT-3, and ChatGPT as the reasoner models for conducting experiments in the fine-tuning, few-shot, and zero-shot settings. Detailed introductions and parameter comparisons for each model are provided in Appendix~\ref{appendix:reasoner-details}, along with implementation details for each model.
We performed a hyper-parameter search for the number of retrieved triples $k$ in $\{5, 10, 20, 30\}$ using TaPEX-Large on the development set of \ourdata. Finally, we chose $k=20$ based on the F1 score. 

\subsection{Main Results}

\begin{table*}[htbp]
  \centering
  \small
    \begin{tabular}{lcccccccccccc}
    \toprule 
    \multicolumn{1}{c}{\multirow{3}[6]{*}{\textbf{Model}}} & \multicolumn{4}{c}{\textbf{Table Only}} & \multicolumn{4}{c}{\textbf{Knowledge Enhanced}} & \multicolumn{4}{c}{\textbf{$\Delta$}} \\
\cmidrule{2-13}          & \multicolumn{2}{c}{\textbf{Dev}} & \multicolumn{2}{c}{\textbf{Test}} & \multicolumn{2}{c}{\textbf{Dev}} & \multicolumn{2}{c}{\textbf{Test}} & \multicolumn{2}{c}{\textbf{Dev}} & \multicolumn{2}{c}{\textbf{Test}} \\
\cmidrule{2-13}          & \textbf{EM} & \textbf{F1} & \textbf{EM} & \textbf{F1} & \textbf{EM} & \textbf{F1} & \textbf{EM} & \textbf{F1} & \textbf{EM} & \textbf{F1} & \textbf{EM} & \textbf{F1} \\
    \midrule
    \textit{Fine-Tuning} &       &       &       &       &       &       &       &       &       &       &       &  \\
    $\text{\tapex}_{\text{large}}$ & 14.44 & 18.52 & 12.83 & 17.1  & \textbf{60.62} & \textbf{63.22} & 56.63 & 58.75 & \cellcolor[rgb]{ .973,  .412,  .42} 46.18 & \cellcolor[rgb]{ .976,  .439,  .447} 44.7 & \cellcolor[rgb]{ .976,  .455,  .463} 43.8 & \cellcolor[rgb]{ .976,  .49,  .498} 41.65 \\
    $\text{BART}_{\text{large}}$ & 9.34  & 13.41 & 8.17  & 12.17 & 51.7  & 54.49 & 52.81 & 56.16 & \cellcolor[rgb]{ .976,  .478,  .486} 42.36 & \cellcolor[rgb]{ .976,  .498,  .506} 41.08 & \cellcolor[rgb]{ .976,  .439,  .447} 44.64 & \cellcolor[rgb]{ .976,  .451,  .459} 43.99 \\
    $\text{BART}_{\text{base}}$ & 7.64  & 11.57 & 8.38  & 11.56 & 45.65 & 48.89 & 46.87 & 50.28 & \cellcolor[rgb]{ .976,  .549,  .561} 38.01 & \cellcolor[rgb]{ .98,  .561,  .569} 37.32 & \cellcolor[rgb]{ .976,  .541,  .549} 38.49 & \cellcolor[rgb]{ .976,  .537,  .545} 38.72 \\
    $\text{T5}_{\text{base}}$ & 9.77  & 14.05 & 9.12  & 12.97 & 45.54 & 48.94 & 46.02 & 49    & \cellcolor[rgb]{ .98,  .588,  .596} 35.77 & \cellcolor[rgb]{ .98,  .604,  .612} 34.89 & \cellcolor[rgb]{ .98,  .569,  .576} 36.9 & \cellcolor[rgb]{ .98,  .584,  .592} 36.03 \\
    \midrule
    \textit{Zero-Shot} &       &       &       &       &       &       &       &       &       &       &       &  \\
    GPT-3 & 8.07  & 17.85 & 10.07 & 20.11 & 33.55 & 45.04 & 36.69 & 47.76 & \cellcolor[rgb]{ .984,  .761,  .769} 25.48 & \cellcolor[rgb]{ .984,  .729,  .741} 27.19 & \cellcolor[rgb]{ .984,  .741,  .749} 26.62 & \cellcolor[rgb]{ .984,  .722,  .733} 27.65 \\
    ChatGPT & 3.82  & 7.65  & 4.03  & 7.31  & 17.73 & 27.53 & 15.69 & 26.79 & \cellcolor[rgb]{ .988,  .953,  .965} 13.91 & \cellcolor[rgb]{ .988,  .851,  .863} 19.88 & \cellcolor[rgb]{ .988,  .988,  1} 11.66 & \cellcolor[rgb]{ .988,  .859,  .871} 19.48 \\
    \midrule
    \textit{Few-Shot} &       &       &       &       &       &       &       &       &       &       &       &  \\
    GPT-3 & \textbf{33.86} & \textbf{39.58} & \textbf{31.81} & \textbf{37.06} & 57.86 & 63.04 & \textbf{60.23} & \textbf{64.89} & \cellcolor[rgb]{ .984,  .784,  .796} 24 & \cellcolor[rgb]{ .984,  .792,  .804} 23.46 & \cellcolor[rgb]{ .984,  .71,  .722} 28.42 & \cellcolor[rgb]{ .984,  .722,  .729} 27.83 \\
    ChatGPT & 20.7  & 23.95 & 19.72 & 23.92 & 45.01 & 49.51 & 43.26 & 49.53 & \cellcolor[rgb]{ .984,  .78,  .788} 24.31 & \cellcolor[rgb]{ .984,  .757,  .769} 25.56 & \cellcolor[rgb]{ .984,  .792,  .804} 23.54 & \cellcolor[rgb]{ .984,  .757,  .769} 25.61 \\
    \bottomrule
    \end{tabular}%
    \caption{Performance of different reasoners on \ourdata. Block $\Delta$ represents the increase in performance after incorporating knowledge base as an additional source of information. We employ the text-davinci-003 version for GPT-3 and the gpt-3.5-turbo version for ChatGPT.}
  \label{tab:results-question-answering}%
\vspace{-0.3cm}
\end{table*}%

As is shown in Table~\ref{tab:results-question-answering}, the inclusion of the knowledge base consistently and significantly enhanced the performance across all experimental settings and model types, as reflected by relative improvements of 1.9 to 6.5 times in EM scores and 1.8 to 4.6 times in F1 scores. These findings highlight the effectiveness of leveraging a knowledge base for question-answering tasks and its potential for improving the accuracy and capability of reasoning systems. More detailed experimental results can be found in Appendix~\ref{appendix:detailed-exp-res}.
We also observed that in the table-only scenario, few-shot GPT-3 outperformed the fine-tuned models, indicating that the LLM itself might possess some stored external knowledge. This issue will be further investigated in Section~\ref{sec:knowledge-source}.

\subsection{Comparison of Knowledge Sources}
\label{sec:knowledge-source}
We performed in-depth experiments to compare two distinct external knowledge sources with KB: \textbf{(i)} LLM-generated Knowledge; \textbf{(ii)} Wikipedia Passages. The collection and utilization of these two types of knowledge are provided in Appendix~\ref{appendix:other-knowledge-sources}.

As is shown in Table~\ref{tab:comparison-knowledge-source}, incorporating a KB as an external source of knowledge can significantly surpass the other two methods in terms of EM and F1 scores. 
We perceive a structured KB to possess several advantages over other forms of knowledge sources:
\textbf{(i)} It provides structured data and semantic relationships, yielding more precise and consistent knowledge in comparison to unstructured text. 
\textbf{(ii)} It ensures accuracy and reliability as the information is validated and sourced from trusted authorities, unlike unstructured text that may contain ambiguity and inconsistencies.
\textbf{(iii)} The structured representation of it enhances the interpretability and comprehension of the source, context, and relationships, in contrast to the relatively opaque nature of LLMs trained on unstructured text. 
\textbf{(iv)} The semantic relationships between entities can aid the model in comprehending the structure of the table.

Furthermore, due to the inclusion of fine-grained gold evidence annotations in \ourdata, we can employ a supervised approach to train a retriever. However, when it comes to Wikipedia passages, only an unsupervised or weakly supervised retriever can be utilized. The discrepancy between training methods of retrievers is equally responsible for the performance disparity.

\begin{table}[htbp]
  \centering
  \small
    \begin{tabular}{lcccc}
    \toprule
    \multicolumn{1}{c}{\multirow{2}[4]{*}{\textbf{Knowledge Source}}} & \multicolumn{2}{c}{\textbf{Dev}} & \multicolumn{2}{c}{\textbf{Test}} \\
\cmidrule{2-5}          & \textbf{EM} & \textbf{F1} & \textbf{EM} & \textbf{F1} \\
    \midrule
    Table Only & 14.44 & 18.52 & 12.83 & 17.1 \\
    LLM & 29.62 & 34.23 & 26.51 & 30.58 \\
    Wikipedia Passages& 32.27 & 36.69 & 28.31 & 32.24 \\
    \tblgray
    Knowledge Base & \textbf{60.62} & \textbf{63.22} & \textbf{56.63} & \textbf{58.75} \\
    \bottomrule
    \end{tabular}%
    \caption{Experimental results with various external knowledge sources. 
    We employed $\text{\tapex}_{\text{large}}$ as a representative reasoner.}
  \label{tab:comparison-knowledge-source}%
\vspace{-0.3cm}
\end{table}%


\subsection{Error Analysis}
For error analysis, we randomly select 100 error cases of the few-shot GPT-3 model from \ourdata development set.
The results of the error analysis indicate that the majority of errors are attributed to: \textbf{(i)} the retriever failing to retrieve relevant external knowledge (39\%); \textbf{(ii)} the reasoner's inadequate understanding of the retrieved knowledge (42\%). Therefore, it is necessary to enhance the retriever's ability to retrieve more accurate external knowledge and improve the reasoner's capability to handle heterogeneous information. More details can be found in Appendix~\ref{appendix:error-analysis}.


\section{Related Work}


\noindent \textbf{Knowledge Enhanced Models} 
A flurry of QA systems involves the use of multiple sources of knowledge to answer a wider range of questions and achieve superior performance on benchmarks that are constructed based on a single knowledge source~\cite{oguz2020unik, he2023anameta, zhen2022survey}.
Available knowledge sources can be divided into three groups: 
\textbf{(i)} unstructured text;
\textbf{(ii)} structured knowledge bases;
\textbf{(iii)} semi-structured tables.
KG-FiD~\cite{DonghanYu2021KGFiDIK} infused knowledge graph in FiD~\cite{izacard2020leveraging} model for ODQA via constructing a graph structure with KB triples and passages.~\citet{chen-etal-2020-hybridqa, zhu2021tat} propose to integrate both tabular and textual content to answer questions. Unik-QA~\cite{oguz2020unik} unified representations of KB triples and semi-structured tables into unstructured text and verify if adding tables as inputs can improve KBQA or TextQA tasks.



\noindent \textbf{QA over Heterogeneous Information}
Reasoning over heterogeneous information poses significant challenges.
Recent works have demonstrated the potential of a single transformer-based model to fuse heterogeneous information ~\cite{xie2022unifiedskg, liu2023zero, liu-etal-2022-uni}. These works often unify the representation of different types of information by reducing them to text.
Another line of work involves using different models to process data from different structures, e.g., graph neural network for knowledge graph~\cite{DonghanYu2021KGFiDIK, gao2019hybrid}. However, the intricate nature of the diverse designs renders it less convenient compared to a transformer-based approach.


\section{Conclusion}

We present \ourdata, the first TableQA dataset that incorporates a knowledge base as supplementary information to tables. 
Significantly, we provide fine-grained gold evidence annotations to facilitate deeper research into the missing knowledge problem of tables resulting from the highly condensed structure.
In our experiments, we devise a retriever-reasoner framework to integrate the knowledge base into TableQA effectively. 
We firmly believe that \ourdata presents an intriguing and demanding challenge for the community to tackle.

\section*{Limitations}
As is shown in Table~\ref{tab:comparison}, \ourdata is not large-scale enough compared to other existing datasets due to the complexity of the labeling task. 
Furthermore, when extracting the sub-graph from Wikidata, we only considered one-hop connections. However, if we extract sub-graphs with more hops, it not only provides a wider range of external knowledge for tables but also presents greater challenges for the reasoner due to the inclusion of more complex structures.
We consider the aforementioned limitations to be areas for future improvement and development of the \ourdata.
In experiments, we have not provided experiment results of more advanced versions of GPT such as GPT4 due to the access and API frequency limitations.



\section*{Ethics Statement}
This paper introduces \ourdata, an openly accessible English dataset designed for the research community to investigate table question-answering.
The annotators we employ possess bachelor's degrees in computer science and are compensated at an hourly wage of \$9, which exceeds the local average salary of similar jobs.
\ourdata is built on \hybridqa~\cite{chen-etal-2020-hybridqa}, which is under the MIT license. The knowledge base is constructed on Wikidata~\cite{DennyVrandecic2014WikidataAF}, which is under the CC0 public domain license. Both allow us to modify and create new datasets.


\bibliography{anthology,custom,2022.emnlp-main}
\bibliographystyle{acl_natbib}

\clearpage
\newpage
\appendix

\section{Probabilistic Formalization of the Model}
\label{appendix:probabilistic-formalization}
As discussed in Section~\ref{sec:task_definition}, the task aims to maximize the probability distribution $ p(a | T, \mathcal{G}, q) $. The whole $\mathcal{G}$ is computationally expensive to process and may introduce noises, as it may contain information that is not related to the specific question. So we retrieve a sub-KB $ \mathcal{G}' $ as the evidence for answering the question instead of directly reasoning on $ \mathcal{G} $. Considering $ \mathcal{G}' $ as latent variables, we rewrite $ p(a | T, \mathcal{G}, q) $ as follows:

\vspace{-0.2cm}
\begin{equation}
{\small
\begin{aligned}
 p(a | T, \mathcal{G}, q)=\sum_{\mathcal{G}'} p_{\theta}(a|T, \mathcal{G}', q) & p_{\beta}(\mathcal{G}'|T, q)
\end{aligned}}
\nonumber
\vspace{-0.2cm}
\label{eq:joint_prob}
\end{equation}

As is indicated in the above equation, the target distribution is jointly modeled by a knowledge base retriever $p_{\beta}(\mathcal{G}'|T, q)$, and a reasoner conditioned on the table and retrieved triples $p_{\theta}(a|T, \mathcal{G}', q)$. The goal of training is to find the optimal parameters $\beta$ and $\theta$ which can maximize the log-likelihood of training data.

\vspace{-0.2cm}
\begin{equation}
{\small
\begin{aligned}
\mathcal{L}(\beta, \theta) = \max_{\beta, \theta} \sum_{\mathcal{D}}\log \sum_{ \mathcal{G}'} p_{\theta}(a|T, \mathcal{G}', q) p_{\beta}(\mathcal{G}'|T, q)
\end{aligned}}
\nonumber
\vspace{-0.2cm}
\label{eq:objective}
\end{equation}

We decouple the two models and train them separately, i.e., we first train the retriever $p_{\beta}$ and then train the reasoner $p_{\theta}$ on the sub-graph sampled by the retriever~\cite{zhang-etal-2022-subgraph, sachan-etal-2021-end}. The above equation can be approximated as follows:

\vspace{-0.2cm}
\begin{equation}
{\small
\begin{aligned}
\mathcal{L}(\beta, \theta) = \max_{\beta, \theta} \sum_{\mathcal{D}} \log p_{\theta}(a|T, \mathcal{G}', q) + \log p_{\beta}(\mathcal{G}'|T, q)
\end{aligned}}
\nonumber
\vspace{-0.2cm}
\label{eq:approximateobjective}
\end{equation}

\section{More Experimental Results}

\subsection{Corresponding Validation Results of Retrievers}
\label{sec:validation-results}

The corresponding validation results of comparison between different retrieval methods are shown in Table~\ref{tab:retriever-main-results-validation-results}.

\begin{table}[bhtp]
\small 
\begin{tabularx}{\linewidth}{lcccc} 
\toprule
& \multicolumn{1}{c}{\textbf{Top-1}} & \multicolumn{1}{c}{\textbf{Top-5}} & \multicolumn{1}{c}{\textbf{Top-20}} & \multicolumn{1}{c}{\textbf{Top-100}} \\
\midrule
Random & 0.00 & 0.61 & 2.38 & 13.50 \\
String Match & 5.65 & 15.93 & 26.49 & 40.83 \\
Cross-Encoder & 36.73 & 60.89 & 80.39 & \textbf{93.10} \\
Bi-Encoder & 25.98 & 48.74 & 70.89 & 88.55 \\
\tblgray
\retriever & \textbf{37.58} & \textbf{63.10} & \textbf{80.99} & 91.66 \\
\bottomrule
\end{tabularx}
\caption{Corresponding validation results of Table~\ref{tab:retriever-main-results}}
\label{tab:retriever-main-results-validation-results}
\end{table}

\subsection{QA Results over Answer Distribution}
\label{appendix:detailed-exp-res}
We divided the dataset based on the answer source and reported the results on the dev set of \ourdata for each subset in Table~\ref{tab:experimental-answer-distribution}. 
After incorporating knowledge, the model's performance shows the greatest improvement in the subset where the answer source is In-KB. 
However, even after incorporating knowledge, the model's performance on calculated answers remains lower compared to the other two categories. This indicates that the model's ability to perform calculations on heterogeneous data still needs improvement.

\begin{table*}[htbp]
  \centering
  \small
    \begin{tabular}{lcccccccccccc}
    \toprule
    \multicolumn{1}{c}{\multirow{3}[6]{*}{\textbf{Model}}} & \multicolumn{6}{c}{\textbf{Table Only}}       & \multicolumn{6}{c}{\textbf{Knowledge Enhanced}} \\
\cmidrule{2-13}          & \multicolumn{2}{c}{\textbf{In Table}} & \multicolumn{2}{c}{\textbf{In-KB}} & \multicolumn{2}{c}{\textbf{Calculated}} & \multicolumn{2}{c}{\textbf{In-Table}} & \multicolumn{2}{c}{\textbf{In-KB}} & \multicolumn{2}{c}{\textbf{Calculated}} \\
\cmidrule{2-13}          & \textbf{EM} & \textbf{F1} & \textbf{EM} & \textbf{F1} & \textbf{EM} & \textbf{F1} & \textbf{EM} & \textbf{F1} & \textbf{EM} & \textbf{F1} & \textbf{EM} & \textbf{F1} \\
    \midrule
    \textit{Fine-Tuning} &       &       &       &       &       &       &       &       &       &       &       &  \\
    $\text{\tapex}_{\text{large}}$ & 22.47 & 27.38 & 8.04  & 11.59 & \textbf{36.36} & \textbf{36.36} & 55.56 & 60.47 & \textbf{59.81} & 61.36 & \textbf{45.45} & \textbf{45.45} \\
    $\text{BART}_{\text{large}}$ & 15.91 & 19.72 & 4.11  & 8.46  & 27.27 & 27.27 & 49.24 & 53.46 & 54.02 & 56.11 & 36.36 & 42.42 \\
    $\text{BART}_{\text{base}}$ & 11.62 & 15.77 & 4.3   & 8.13  & 27.27 & 27.27 & 40.15 & 44.68 & 50.09 & 52.42 & 27.27 & 27.27 \\
    $\text{T5}_{\text{base}}$ & 15.91 & 21.9  & 4.86  & 7.98  & 27.27 & 27.27 & 45.71 & 50.83 & 45.61 & 47.68 & 36.36 & 42.42 \\
    \midrule
    \textit{Zero-Shot} &       &       &       &       &       &       &       &       &       &       &       &  \\
    GPT-3 & 13.38 & 23.93 & 4.3   & 13.62 & 0     & 4.48  & 23.48 & 38.56 & 44.68 & 50.53 & 0     & 11.89 \\
    ChatGPT & 7.83  & 14.14 & 0.93  & 2.88  & 0     & 5.58  & 18.69 & 31.05 & 17.38 & 25.42 & 0     & 3.21 \\
    \midrule
    \textit{Few-Shot} &       &       &       &       &       &       &       &       &       &       &       &  \\
    GPT-3 & \textbf{32.83} & \textbf{40.43} & \textbf{35.14} & \textbf{39.53} & 9.09  & 12.73 & \textbf{57.32} & \textbf{64.38} & 58.88 & \textbf{62.75} & 27.27 & 28.93 \\
    ChatGPT & 22.22 & 26.83 & 20    & 22.12 & 0     & 9.7   & 46.97 & 53.89 & 44.49 & 47.11 & 0     & 8.33 \\
    \bottomrule
    \end{tabular}%
    \caption{Experimental results based on answer distribution on dev set of \ourdata.}
  \label{tab:experimental-answer-distribution}%
\end{table*}%





\section{More Details on Dataset}

\subsection{Dataset Preprocessing}
\label{appendix:dataset-details-qa-filtering}
Due to the post-create question-answer annotation approach used in HybridQA, which involves collecting tables and Wikipedia passages and then having annotators label question-answer pairs, there is a significant overlap between some of the questions and the passages from Wikipedia. We consider such examples to be lacking in naturalness. Therefore, we have designed the following rules for filtering:
\textbf{(i)} We eliminated questions with an LCS Similarity of 0.7 or higher, which was calculated by dividing the length of the Longest Common Sub-sequence (LCS) between the question and its corresponding Wikipedia passage by the length of the question.
\textbf{(ii)} We employed a fuzzy matching technique to filter and retain question-answer pairs that have corresponding words in the knowledge base but are not found in the table.

\subsection{Low-Quality Annotation Filtering Rules}
\label{appendix:low-quality-filtering}
We applied several rules to filter low-quality annotations:
\textbf{(i)} The source of the answer is invalid. As mentioned in Section~\ref{sec: question-answer-annotation}, the answer can only originate from three valid sources. During the annotation process, we instructed the annotators to manually indicate the answer source. In the final review, we employ rule-based methods to trace back the answers and verify their sources as a double-checking measure; \textbf{(ii)} Gold evidence was not explicitly marked; \textbf{(iii)} The annotated evidence is invalid. For example, the index exceeds the range of the table or there is no corresponding triple in the sub-graph.

\section{More Details on Retrievers}
\label{appendix:retriever-details}

\subsection{Training of Bi-Encoder}
\label{appendix:training-bi-encoder}
The training process of Bi-Encoder is similar to that in~\cite{karpukhin-etal-2020-dense}. 
We optimize the model using a contrastive learning loss function:
\begin{equation}
\begin{aligned}
&\mathcal{L}(q_i, T_i, t^+_{i,1}, \cdots, t^+_{i,p}, t^-_{i,1}, \cdots, t^-_{i,n}) =\\
& -\sum_{j=1}^m \log \frac{ e^{s(t^+_{i, j}, q_i, T_i)} }{{e^{s(t^+_{i, j}, q_i, T_i)}} + \sum_{k=1}^n{e^{s(t^-_{i, k}, q_i, T_i)}}}
\end{aligned}
\nonumber
\label{eq:retriever-objective}
\end{equation}

\subsection{Implementation Details}
\label{appendix:retriever-implementation-details}
We use two independent BERT networks~\cite{devlin2019bert} (base, uncased) for retrieval bi-encoder and a single RoBERTa~\cite{liu2019roberta} model for re-ranker cross-encoder. 
During the training process, both models are trained on the train set of \ourdata with a learning rate of 10-5 using Adam, linear scheduling with warm-up and dropout rate 0.1. Bi-encoder is trained up to 20 epochs with a batch size of 16, while cross-encoder is trained up to 5 epochs with a batch size of 32.

\subsection{Negative Sampling}
\label{appendix:number-of-negative-examples}

According to Table~\ref{tab:ablation-study-negative-number}, we conducted experiments with varying numbers of negative examples ($N \in \{25, 50, 100\}$). We observed that increasing the number of negative examples does not necessarily lead to improved model performance. For the cross-encoder, the model achieves its highest performance at $N=50$, while for the bi-encoder, it is at $N=25$.

\begin{table}[htbp]
  \centering
    \begin{tabular}{ccccc}
    \toprule
    \textbf{\#N} & \textbf{Top-1} & \textbf{Top-5} & \textbf{Top-20} & \textbf{Top-100} \\
    \midrule
    \multicolumn{5}{l}{\textit{Bi-Encoder}} \\
    \tblgray
    25    & \textbf{29.17} & \textbf{51.95} & \textbf{72.12} & \textbf{89.62} \\
    50    & 25.68 & 48.17 & 68.11 & 87.94 \\
    100   & 22.42 & 39.47 & 62.51 & 85.03 \\
    \midrule
    \multicolumn{5}{l}{\textit{Cross-Encoder}} \\
    25    & 27.53 & 54.32 & 73.76 & 91.63 \\
    \tblgray
    50    & \textbf{37.83} & \textbf{63.84} & \textbf{82.14} & \textbf{94.44} \\
    100   & 12.49 & 28.63 & 44.76 & 70.22 \\
    \bottomrule
    \end{tabular}%
    \caption{Ablation study on the number of negative numbers (\#N). \textcolor{gray}{Gray} represents the final model selection for \retriever}
  \label{tab:ablation-study-negative-number}%
\end{table}%

\section{More Details on Reasoners}
\label{appendix:reasoner-details}

\subsection{Introduction}
\label{appendix:reasoner-introduction}

\noindent \textbf{\tapex}~\cite{liu2021tapex} guides the language model to mimic a SQL executor on the synthetic corpus, resulting in groundbreaking results on four table-related datasets. We take \tapex as a representative of tabular language models.


\noindent \textbf{BART}~\cite{lewis-etal-2020-bart} and \textbf{T5}~\cite{raffel2020t5} are representatives of general pre-trained encoder-decoder language models, and have performed exceptionally well on a wide range of NLP tasks.

\noindent \textbf{GPT-3}~\cite{gpt3} is one of the top-performing models among the large language models with a decoder architecture. It exhibits strong question answering capabilities in both zero-shot and few-shot settings.


\noindent \textbf{ChatGPT} is a variant of GPT-3, which is trained using Reinforcement Learning from Human Feedback (RLHF). It excels in natural language conversations and exhibits human-like responses.

Table~\ref{tab:model-parameter} shows the comparison of parameters of different models.

\subsection{Evaluation Metrics}
\label{appendix:evaluation-metrics-qa}
\noindent \textbf{Exact Match} The EM score is a strict all-or-nothing metric, which represents the percentage of predictions that exactly match the ground truth.

\noindent \textbf{F1 Score} is another widely-used metric in QA~\cite{chen-etal-2020-hybridqa, zhu2021tat}, which measures the token overlap between the predicted answer and ground truth.

\subsection{Implementation Details}

The experimental settings for fine-tuning models include using the AdamW optimizer with an initial learning rate of 5e-5, training for 20 epochs, and using a batch size of 24. The training process takes about 7.8 hours for BART-Large/TaPEX-Large and 5.2 hours for BART-Base/T5-Base with 4 16G V100 GPUs.
The few-shot model utilized the KATE~\cite{liu-etal-2022-makes} method, where for each sample, five examples were retrieved from the training set. 
For both the zero-shot model and the few-shot, we employed a greedy decoding strategy ($t=0$) to obtain the final answer. The inference process typically takes approximately 5.5 seconds per question with OpenAI API.

\begin{table}[htbp]
  \centering
  \small
    \begin{tabular}{ll}
    \toprule
    \textbf{Model} & \textbf{\#Parameter} \\
    \midrule
    $\text{\tapex}_{\text{large}}$ & 400 million \\
    $\text{BART}_{\text{large}}$ & 400 million \\
    $\text{BART}_{\text{base}}$ & 140 million \\
    $\text{T5}_{\text{base}}$ & 220 million \\
    GPT-3 & 175 billion \\
    \bottomrule
    \end{tabular}%
    \caption{Parameter of reasoners}
  \label{tab:model-parameter}%
\end{table}%



\section{Other External Knowledge Sources}
\label{appendix:other-knowledge-sources}
\noindent \textbf{LLM-generated Knowledge} Numerous studies have proposed utilizing Large Language Models (LLMs) as databases~\cite{liu-etal-2022-generated, yu2023generate}. We employed the prompt "\textit{Generate some knowledge about the given question and table}" to instruct the LLM (text-davinci-003 in our case) to generate knowledge that is beneficial for answering the current question with a greedy decoding strategy.

\noindent \textbf{Wikipedia Passage} As discussed in~\cite{chen-etal-2020-hybridqa}, hyperlinked Wikipedia passages can provide additional information that complements the table. Following the methodology outlined in~\cite{chen-etal-2020-hybridqa}, we employed the same passage retriever to retrieve relevant knowledge for the current question.

We concatenated the question, the serialized table, and the external knowledge as the input of PLMs and take the output as the final answer, which is the same as in Section~\ref{sec:model-reasoner}.

\section{Detailed Error Analysis Result}
\label{appendix:error-analysis}
Errors are categorized into four types: 
(1) \textbf{Knowledge Uncovered}: The provided knowledge does not include the necessary information to answer the current question.
(2) \textbf{Erroneous Knowledge}: The provided knowledge is detrimental to answering the current question or is factually incorrect.
(3) \textbf{Reasoning Error}: The provided knowledge can answer the current question, but the reasoner failed to provide the correct response.
(4) \textbf{False Negative}: Misjudged by the evaluator.
As is shown in Figure~\ref{fig:error-type}, the reasoning capability of the reasoner needs to be further enhanced in handling heterogeneous data sources, as issues resulting from the incorrect processing of information by the reasoner accounted for the highest proportion of errors, reaching 42\%. 
Additionally, a considerable percentage (39\%) is attributed to the absence of necessary information in the provided knowledge, highlighting the need for an improved retriever to offer more precise and comprehensive knowledge. 

\section{Question Types}
\label{sec:question-types-other-datasets}

\begin{figure*}[htbp]
\vspace{-10pt}
    \centering
    \begin{subfigure}[b]{0.49\textwidth}
        \centering       
        \includegraphics[width=0.99\textwidth]{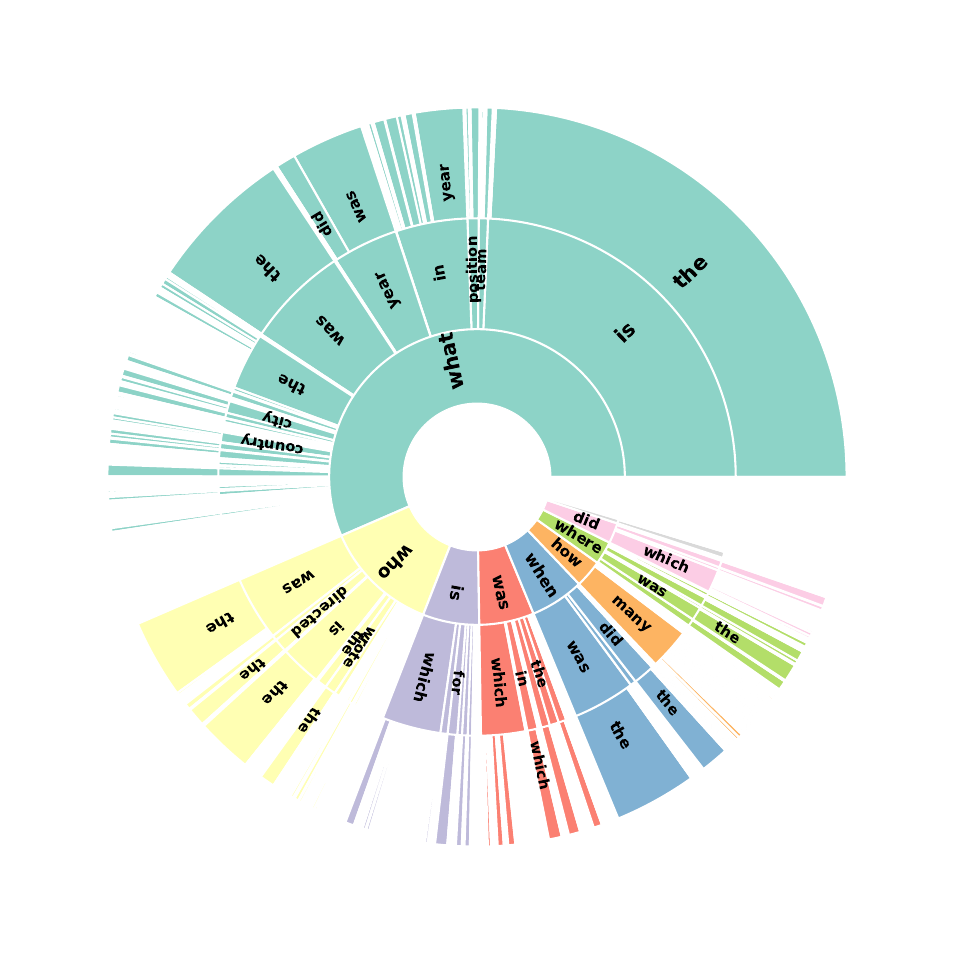}
        \caption{Distribution of question types in \ourdata.}
        \label{fig:question-types}
    \end{subfigure}
    \hfill
    \begin{subfigure}[b]{0.49\textwidth}
        \centering
        \includegraphics[width=0.99\textwidth]{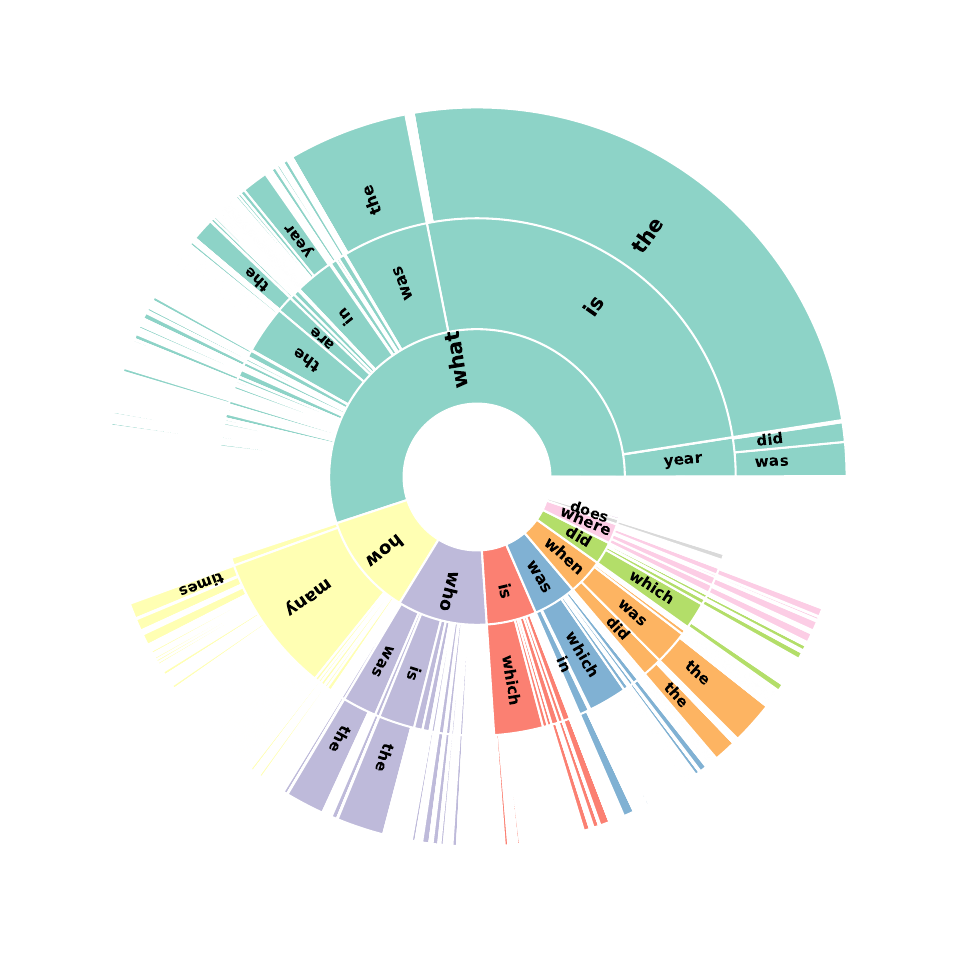}
        \caption{Distribution of question types in \hybridqa. }
        \label{fig:question-types-hybridqa}
    \end{subfigure}
    \caption{Question types of two datasets. Question types are extracted using methods proposed in~\cite{yang-etal-2018-hotpotqa}, which utilize heuristics based on identifying the central question word (CQW) and analyzing surrounding tokens.}
    \label{fig:qt}
\vspace{-0.2cm}
\end{figure*}

We visualized the question types in \ourdata using the same method as in~\cite{chen-etal-2020-hybridqa, yang-etal-2018-hotpotqa}. Question types of \ourdata and \hybridqa are shown in Figure~\ref{fig:question-types} and Figure~\ref{fig:question-types-hybridqa}, respectively. Although \ourdata is built on HybridQA, it has a higher percentage of \textit{who} questions. 
Moreover, in \ourdata, there is a higher proportion of questions that include the terms "position," "team," "city," and "country." 

\end{document}